\documentclass[letterpaper, 10 pt, conference]{ieeeconf}  

\IEEEoverridecommandlockouts                              

\usepackage{cite}
\usepackage{amsmath, bm}
\usepackage{amssymb,amsfonts}
\usepackage{algorithm}
\usepackage{algpseudocode}
\usepackage{array}
\usepackage{graphicx}
\usepackage{textcomp}
\usepackage{float}
\usepackage{xcolor}
\usepackage{csquotes}
\usepackage{array}
\usepackage{makecell}
\usepackage[pscoord]{eso-pic}
\DeclareMathOperator{\E}{\mathbb{E}}
\DeclareMathOperator*{\argmax}{arg\,max}

\newcommand{\placetextbox}[3]{
 \setbox0=\hbox{#3}
 \AddToShipoutPictureFG*{ \put(\LenToUnit{#1\paperwidth},\LenToUnit{#2\paperheight}){\vtop{{\null}\makebox[0pt][c]{#3}}}
 }
 }
 \placetextbox{.5}{0.055}{\small{Accepted at the Workshop on Safe and Reliable Robot Autonomy under Uncertainty at ICRA 2022, Philadelphia, USA}}

\title{\LARGE \bf
Uncertainty Quantification for Competency Assessment \\ of Autonomous Agents}

\author{Aastha Acharya$^{1, 2}$, Rebecca Russell$^{2}$ and Nisar R. Ahmed$^{1}$
\thanks{The authors are with $^{1}$Ann and H.J. Smead Department of Aerospace Engineering Sciences at the University of Colorado Boulder, Boulder, Colorado and $^{2}$The Charles Stark Draper Laboratory, Inc., Cambridge, Massachusetts. {\tt\small aastha.acharya@colorado.edu, rrussell@draper.com, nisar.ahmed@colorado.edu}}
}

\begin{document}

\maketitle
\thispagestyle{empty}
\pagestyle{empty}

\begin{abstract}

For safe and reliable deployment in the real world, autonomous agents must elicit appropriate levels of trust from human users. One method to build trust is to have agents assess and communicate their own competencies for performing given tasks. Competency depends on the uncertainties affecting the agent, making accurate uncertainty quantification vital for competency assessment. In this work, we show how ensembles of deep generative models can be used to quantify the agent's aleatoric and epistemic uncertainties when forecasting task outcomes as part of competency assessment.

\end{abstract}

\section{INTRODUCTION}

While real world deployment of autonomous robotic agents will greatly benefit humans, the challenge lies in establishing appropriate levels of trust that lead to proper usage of and reliance on the autonomy \cite{trust_in_automation, buildingtrust}. Trust is impacted by the human belief in the safety and reliability of the autonomous agents. One method of establishing trust is to have the autonomous robot report its own \emph{competency} by providing self-assessment of its capability to perform the given tasks under varying environmental and dynamical factors \cite{israelsen,Aitken2016AssurancesAM}. Our approach to competency assessment relies on the autonomous agent performing accurate long-horizon forecasting of task outcomes \cite{acharya2022competency}. A critical component of this, particularly for deep reinforcement learning (RL) based autonomous agents, is uncertainty quantification. 

The uncertainties encountered by a deep RL autonomous agent can be categorized as aleatoric or epistemic \cite{uncgen, kendall_gal}. Aleatoric uncertainty is an irreducible form of uncertainty reflected in the training data of the deep learning model. Its sources include process/measurement noise and any unobserved or partially observed variables. Epistemic uncertainty is a reducible form of uncertainty that reflects uncertainty in the model parameters. Its sources include insufficient or out-of-distribution data and any modelling errors. Both types of uncertainties are encountered in the real world when operating in an inherently stochastic environment and/or when the agent is tasked in scenarios outside of its training data distribution. Thus, quantification of both uncertainties is vital for the agent's competency assessment.

In this work, we show preliminary methods to quantify aleatoric and epistemic uncertainties for a deep model-based RL autonomous agent using an ensemble of deep generative models. We present our results for three environmental variations (deterministic, stochastic and out-of-distribution) and show how the two types of uncertainties evolve in each. By presenting the uncertainty quantification results alongside the autonomous agent's task outcome forecast, we show how an informative competency assessment can be generated. 

\section{METHODS}\label{methods}

\begin{figure*}[hbt]
  \centering
  \includegraphics[scale=0.49]{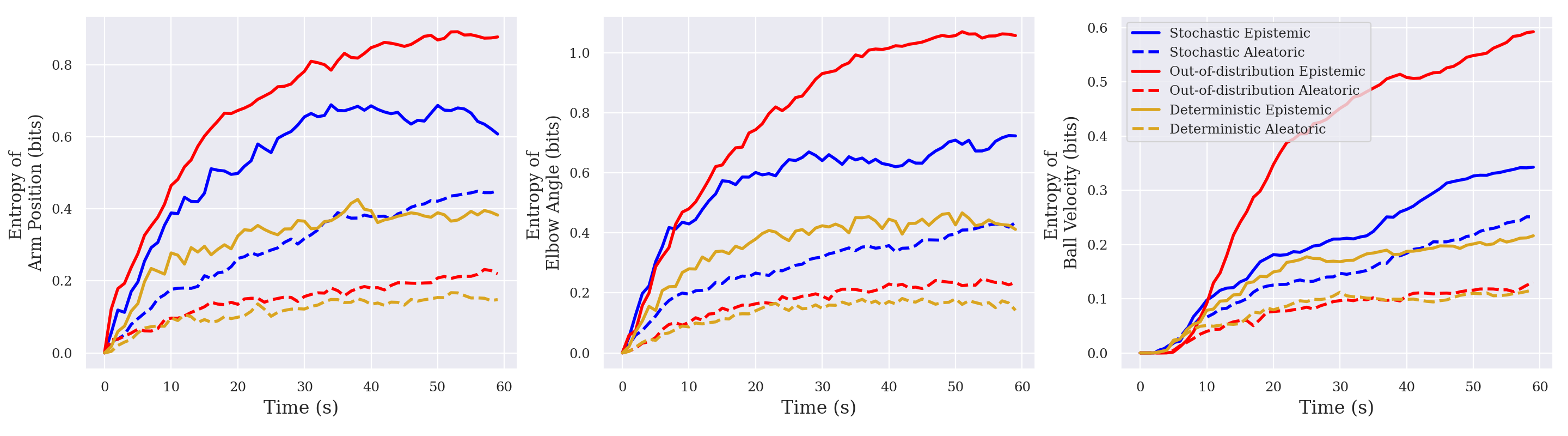}
  \caption{Uncertainty results, represented as entropy values, for Pusher robot under varying environmental conditions. (Left) Uncertainties associated with the position of the arm of the robot. (Middle) Uncertainties associated with elbow angle. (Right) Uncertainties associated with the velocity of the ball. }
  \label{uncertainty_plot}
\end{figure*}

\subsection{Deep Generative Models for Aleatoric Uncertainty} \label{aleatoric_unc}

Generative models can approximate the joint probability distribution underlying training data. We use a conditional VAE that uses the latent space to capture the stochasticity of the data \cite{Kingma2014, condVAE}. By adding a recurrent structure to the VAE, the temporal correlations of the data can also be captured. This is an important feature to learn the dynamics of model-based RL agents, which allows long-horizon forecasting of task outcomes for competency assessment. Given a trajectory of the form $\mathcal{\bm{T}} = \{\bm{s}_0, \bm{a}_0, \bm{s}_1, \bm{a}_1, \ldots, \bm{s}_t, \bm{a}_{t}, \ldots\}$, where $\bm{s}_t$ and $\bm{a}_t$ represent the state and action at time $t$, the recurrent conditional VAE learns $p_\theta(\bm{s}_{1:T}|\bm{s}_0; \bm{a}_{0:T-1})$, which is an approximate distribution under model parameters $\theta$ of states up to time horizon $T$ given only the initial state and sequence of actions. Since aleatoric uncertainty emerges from the training data distribution, the distribution learned by the VAE represents an approximation to true aleatoric uncertainty $p(\bm{s}_{0:T}|\bm{s}_0; \bm{a}_{0:T-1})$. 

\subsection{Deep Ensembles for Epistemic Uncertainty}

The deep generative model from Section \ref{aleatoric_unc} is trained to learn model parameters $\theta^*$ such that 
\begin{equation}
    \theta^* = \argmax_\theta \prod^n_{i=1} p_\theta(\bm{s}^i_{1:T}|\bm{s}^i_0; \bm{a}^i_{0:T-1})
\end{equation}
\noindent where $n$ is the number of training samples. Uncertainties associated with the learned parameters $\theta$, which also impact the approximation of aleatoric uncertainty, are accounted for using epistemic uncertainty. Epistemic uncertainty reflects the amount of training a model has and increases for out-of-distribution data. In our case, this occurs when the trained model is queried with initial state $\bm{s}_0$ and/or sequence of actions $\bm{a}_{0:T-1}$ that are out of its training data distribution. Quantification of epistemic uncertainty relies on having a posterior distribution over the model parameters, which we achieve by training an ensemble of recurrent VAEs \cite{ensembles}.  

\subsection{Uncertainty Quantification}

Given that we are working with a conditional generative model, existing methods for quantifying predictive uncertainty can be used for quantifying and decomposing the two forms of uncertainty \cite{depeweg2017decomposition, notin2020principled}. Following these methods, the separation between the epistemic ($\mathcal{U}_{epistemic}$) and aleatoric ($\mathcal{U}_{aleatoric}$) uncertainties is presented as follows: 

\vspace*{-\baselineskip}

\begin{multline}\label{epistemic}
    \mathcal{U}_{epistemic} = \mathcal{H}(p(\bm{s}_{1:T}|\bm{s}_0; \bm{a}_{0:T-1})) 
    \\
    - \E_{p(\theta|\mathcal{D})}[\mathcal{H}(p(\bm{s}_{1:T}|\bm{s}_0; \bm{a}_{0:T-1}; \theta)) 
\end{multline}

\vspace*{-\baselineskip}

\begin{equation}\label{aleatoric}
    \mathcal{U}_{aleatoric} = \E_{p(\theta|\mathcal{D})}[\mathcal{H}(p(\bm{s}_{1:T}|\bm{s}_0; \bm{a}_{0:T-1}; \theta))]
\end{equation}

\noindent where $\mathcal{H}(\cdot)$ represents the entropy function and $\E_{p(\theta|\mathcal{D})}$ represents the expectation under the posterior distribution of the model parameters $\theta$ given the training data $\mathcal{D}$. 

\section{PRELIMINARY RESULTS}

In this section, we present preliminary results for the Pusher robot, a modified OpenAI gym \cite{brockman2016openai} Reacher environment with a ball added to induce interesting agent behavior. In addition to the standard deterministic environment, we add complexity during training and testing in two ways: 

\begin{itemize}
    \item \textbf{Stochastic training: } Gaussian noise is injected into the system via the actions, and the model is trained on this uncertain environment.
    \item \textbf{Out-of-distribution testing: } Trained model is queried using states that are out of range from the training distribution for the standard deterministic environment. 
\end{itemize}

The uncertainty entropies over time, calculated from Monte Carlo runs using all available ensemble models, are shown in Figure \ref{uncertainty_plot} for three selected state dimensions. In general, epistemic uncertainty value tends to be higher than aleatoric, regardless of environmental condition. As expected, epistemic uncertainty is highest for out-of-distribution condition and aleatoric uncertainty is highest for the stochastic condition. For the out-of-distribution case, since we maintain deterministic conditions, the aleatoric uncertainty is comparable to that of the deterministic environment. We also present corresponding numerical entropy values averaged over time in Table \ref{pusher_ca}, which shows the type of uncertainty that dominates in each scenario. 

To demonstrate the implications of uncertainty on competency assessment, we also present the agent's forecasted outcome probabilities \cite{acharya2022competency} in Table \ref{pusher_ca} when tasked to get the ball to a target location within 30 seconds. The probabilities are assessed based on user specified success-failure boundary for the given task. Here, the agent is least confident in successfully completing the task in the stochastic environment, where it encounters the highest aleatoric uncertainty. This indicates that there is an increased variation in the forecasted outcomes at the success-failure boundary for the stochastic environment. Additional study of the relationships seen in Table \ref{pusher_ca} under different success-failure boundaries and varying out-of-distribution scenarios is our planned future work. Overall, we can clearly see the benefit of assessing and reporting the competency in this form as it can lead to better tasking of the autonomous agent by the human user, thus leading to safe and reliable deployment of autonomy. 

\vspace*{-4pt}

\begin{table}[hbt!]
\caption{Competency Assessment for Pusher Robot} 
\begin{center}
\renewcommand{\arraystretch}{1.2}
\begin{tabular}{ {c}{c}{c}{c}{c} }
 \hline
     & \makecell{Total \\ Uncertainty \\(Bits)} &  \makecell{Total \\ Aleatoric \\(Bits)} & \makecell{Total \\ Epistemic \\(Bits)} & \makecell{Forecasted \\ Probability \\ of Success}
    \\
    \hline 
    \makecell{Deterministic}  & $0.505$ & $0.129$ & $0.376$ & $99.6\%$ \\
    \makecell{Stochastic} & $0.794$ & $0.286$ & $0.508$ & $75.2\%$ \\
    \makecell{Out-of-distribution} & $0.843$ & $0.153$ & $0.690$ & $84.3\%$\\
\hline
\end{tabular}
\end{center}
\label{pusher_ca}
\end{table}

\vspace*{-\baselineskip}

\section{CONCLUSIONS AND FUTURE WORK}

In this work, preliminary results for performing uncertainty quantification for the purpose of competency assessment are presented. We use ensembles of conditional recurrent VAEs to capture and decompose aleatoric and epistemic uncertainties. The results are presented on three different environmental variations to demonstrate the benefits of quantifying the uncertainties in this way. 

We will be expanding this work further by closely studying the ties between the uncertainty types and the agent's performance. In particular, we plan to analyze a wider variety of tasks to understand the general impact of the two uncertainty types on the agent's forecasted performance. Additionally, we will also explore better metrics to more informatively communicate the uncertainties as part of competency assessment. Finally, we plan to study how the two uncertainty forms may be intertwined with each other to ensure that there is no double-counting or miscounting of uncertainty.

\addtolength{\textheight}{-14cm}



\section*{ACKNOWLEDGMENT}

This material is based upon work supported by the Defense Advanced Research Projects Agency (DARPA) under Contract No. HR001120C0032. Any opinions, findings and conclusions or recommendations expressed in this material are those of the author(s) and do not necessarily reflect the views of DARPA.



\bibliographystyle{plain}
\bibliography{ref}

\end{document}